\documentclass[10pt,twocolumn,letterpaper]{article}

\usepackage{cvpr}
\usepackage{times}
\usepackage{epsfig}
\usepackage{graphicx}
\usepackage{amsmath}
\usepackage{amssymb}

\usepackage[utf8]{inputenc}
\usepackage{mathtools}
\usepackage{xcolor}
\usepackage[normalem]{ulem}
\usepackage{etoolbox}
\usepackage{multirow}

\usepackage[pagebackref=true,breaklinks=true,letterpaper=true,colorlinks,bookmarks=false]{hyperref}

\renewcommand{\vec}[1]{\boldsymbol{#1}}
\newcommand{\mat}[1]{\boldsymbol{#1}}

\newcommand{\smpl}[0]{M}
\newcommand{\posefun}[0]{T}
\newcommand{\blendfun}[0]{W}
\newcommand{\offsetfun}[0]{B}
\newcommand{\jointfun}[0]{J}
\newcommand{\jointtransfun}[0]{G}

\newcommand{\pose}[0]{\boldsymbol{\theta}}
\newcommand{\shape}[0]{\boldsymbol{\beta}}

\newcommand{\blendweight}[0]{w}
\newcommand{\blendweights}[0]{\mathbf{W}}
\newcommand{\template}[0]{\mathbf{T}}

\newcommand{\templatetpose}[0]{\mathbf{T}_\mu}

\newcommand{\vertex}[0]{\mathbf{v}}
\newcommand{\ray}[0]{\mathbf{r}}

\newcommand{\vertexweight}[0]{\tau}

\newcommand{\offsetfunvert}[0]{b}

\newcommand{\offsets}{\mathbf{D}}

\cvprfinalcopy 

\def\cvprPaperID{2769} 

\begin{document}

\title{Video Based Reconstruction of 3D People Models\vspace{-2mm}}

\author{Thiemo Alldieck\textsuperscript{1}\hspace{-3mm}
\and
Marcus Magnor\textsuperscript{1}\hspace{-3mm}
\and
Weipeng Xu\textsuperscript{2}\hspace{-3mm}
\and
Christian Theobalt\textsuperscript{2}\hspace{-3mm}
\and
Gerard Pons-Moll\textsuperscript{2}
}

\makeatletter
\let\@oldmaketitle\@maketitle
\renewcommand{\@maketitle}{
	\@oldmaketitle
	\centering
	\vspace{-8mm}
	{\small \textsuperscript{1}Computer Graphics Lab, TU Braunschweig, Germany}\\
	{\small	\textsuperscript{2}Max Planck Institute for Informatics, Saarland Informatics Campus, Germany}\\
	{\tt\scriptsize \{alldieck,magnor\}@cg.cs.tu-bs.de \{wxu,theobalt,gpons\}@mpi-inf.mpg.de}\\
	\vspace{1mm}
	\includegraphics[width=\textwidth]{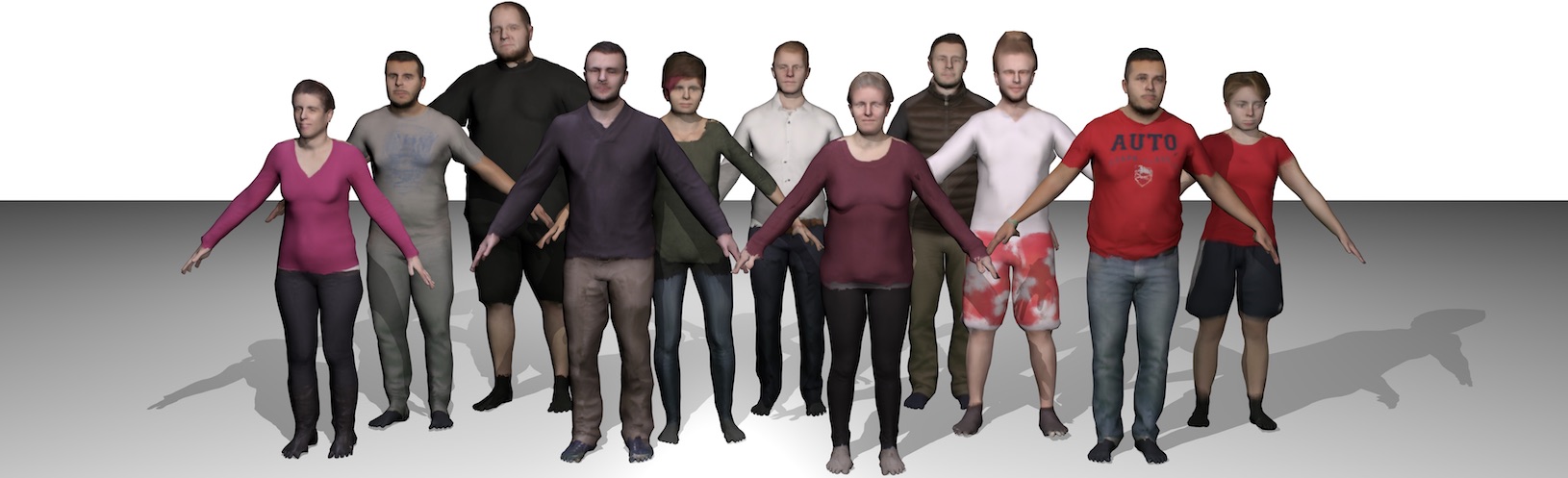}
	\refstepcounter{figure}\normalfont Figure~\thefigure: Our technique allows to extract for the first time accurate 3D human body models, including hair and clothing, from a single video sequence of the person moving in front of the camera such that the person is seen from all sides.
	\label{fig:teaser}
	\vspace{3mm}
}
\makeatother

\maketitle


\begin{abstract}
\vspace{-3mm}
This paper describes a method to obtain accurate 3D body models and texture of arbitrary people from a single, monocular video in which a person is moving.
Based on a parametric body model, we present a robust processing pipeline to infer 3D model shapes including clothed people with 4.5mm reconstruction accuracy.
At the core of our approach is the transformation of dynamic body pose into a canonical frame of reference. Our main contribution is a method to transform the silhouette cones corresponding to dynamic human silhouettes to obtain a visual hull in a common reference frame.
This enables efficient estimation of a consensus 3D shape, texture and implanted animation skeleton based on a large number of frames.
Results on 4 different datasets demonstrate the effectiveness of our approach to produce accurate 3D models. 
Requiring only an RGB camera, our method enables everyone to create their own fully animatable digital double, e.g., for social VR applications or virtual try-on for online fashion shopping.
\end{abstract}
\vspace{-5mm}

\section{Introduction}
\label{sec:introduction}
\begin{figure*}
	\includegraphics[width=0.95\textwidth]{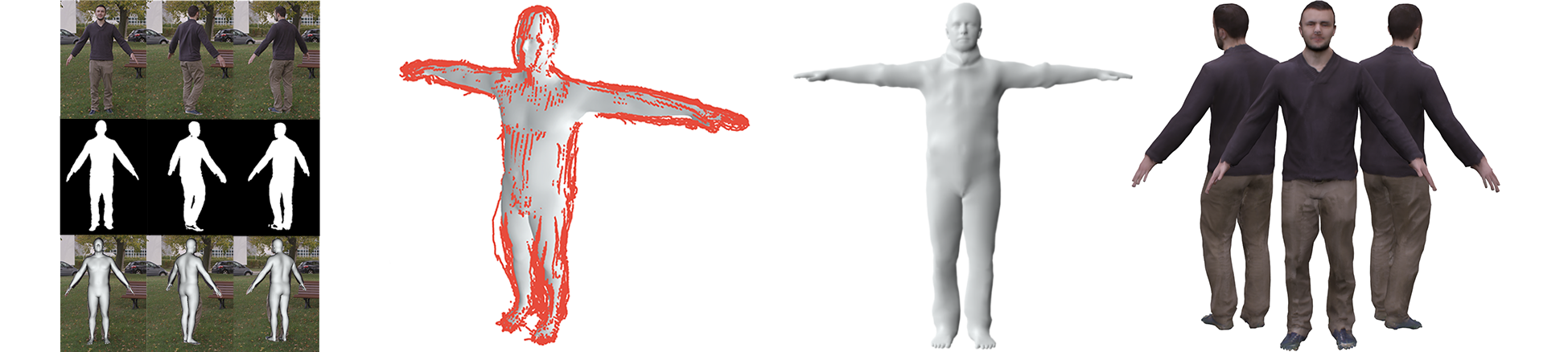}
	{\footnotesize
		\begin{minipage}[t]{0.95\textwidth}
			\begin{minipage}[t]{0.25\textwidth}\centering a)\end{minipage}%
			\begin{minipage}[t]{0.2\textwidth}\centering b)\end{minipage}%
			\begin{minipage}[t]{0.33\textwidth}\centering c)\end{minipage}%
			\begin{minipage}[t]{0.12\textwidth}\centering d)\end{minipage}\\
		\end{minipage}%
	}
	\vspace{-2mm}
	\caption{Overview of our method. The input to our method is an image sequence with corresponding segmentations. We first calculate poses using the SMPL model (a). Then we unpose silhouette camera rays (unposed silhouettes depicted in red) (b) and optimize for the subjects shape in the canonical T-pose (c). Finally, we are able to calculate a texture and generate a personalized blend shape model (d).}
	\label{fig:overview}
	\vspace{-4mm}
\end{figure*}

\vspace{-1mm}
A personalized realistic and animatable 3D model of a human is required for many applications, including virtual and augmented reality, human tracking for surveillance, gaming, or biometrics. This model should comprise the person-specific static geometry of the body, hair and clothing, alongside a coherent surface texture.

One way to capture such models is to use expensive active scanners. 
But size and cost of such scanners prevent their use in consumer applications.
Alternatively, multi-view passive reconstruction from a dense set of static body pose images can be used~\cite{fuhrmann2014mve,Newcombe2011DTAM}. However, it is hard for people to stand still for a long time, and so this process is time-consuming and error-prone. Also, consumer RGB-D cameras can be used to scan 3D body models~\cite{3Dportraits}, but these specialized sensors are not as widely available as video. 
Further, all these methods merely reconstruct surface shape and texture, but no rigged animation skeleton inside.   
All aforementioned applications would benefit from the ability to automatically reconstruct a personalized movable avatar from monocular RGB video. \\
Despite remarkable progress in reconstructing 3D body models~\cite{Bogo:ICCV:2015,weiss2011home,zhang2014quality} or free-form surface~\cite{zollhofer2014real,newcombe2015dynamicfusion,orts2016holoportation,dou2016fusion4d} from depth data, 3D reconstruction of humans in clothing from monocular video (without a pre-recorded scan of the person) has not been addressed before. In this work, we estimate the shape of people in clothing from a single video in which the person moves. 
Some methods infer shape parameters of a parametric body model from a single image~\cite{bogo2016smplify,dibra2017human,bualan2008naked,hasler2010multilinear,zhou2010parametric,jain2010moviereshape}, but the reconstruction is limited to the parametric space and 
can not capture personalized shape detail and clothing geometry.\\ 
To estimate geometry from a video sequence, we could jointly optimize a single free-form shape constrained by a body model to fit a set of $F$ images. Unfortunately, this requires to optimize $F$ poses at once and more importantly it requires storing $F$ models in memory during optimization which makes it computationally expensive and unpractical. 

The key idea of our approach is to generalize visual hull methods~\cite{matusik2000image} to monocular videos of people in motion. Standard visual hull methods capture a static shape from multiple views. Every camera ray through a silhouette point in the image casts a constraint on the 3D body shape. To make visual hulls work for monocular video of a moving person it is necessary to ``undo'' the human motion and bring it to a canonical frame of reference.  
In this work, the geometry of people (in wide or tight clothing) is represented as a deviation from the SMPL parametric body model~\cite{smpl2015loper} of naked people in a canonical T-pose; this model also features a pose-dependent non-rigid surface skinning. 
We first estimate an initial body shape and 3D pose at each frame by fitting the SMPL model to 2D detections similar to~\cite{Lassner,bogo2016smplify}. Given such fits, we associate every silhouette point in every frame to a 3D point in the body model. 
We then transform every projection ray according to the inverse deformation model of its corresponding 3D model point; we call this operation unposing (Fig.~\ref{fig:unpose_rays}). 
After unposing the rays for all frames we obtain a visual hull that constrains the body shape in a canonical T-pose. 
We then jointly optimize body shape parameters and free-form vertex displacements to minimize the distance between 3D model points and unposed rays. This allows us to efficiently optimize a single displacement surface on top of SMPL constrained to fit all frames at once, which requires storing only one model in memory (Fig.~\ref{fig:overview}).  
Our technique allows for the first time extracting accurate 3D human body models, including hair and clothing, from a single video sequence of the person moving in front of the camera such that the person is seen from all sides.\\
Our results on several 3D datasets show that our method can reconstruct 3D human shape to a remarkable accuracy of 4.5~mm (even higher 3.1~mm with ground truth poses) despite monocular depth ambiguities. We provide our dataset and source code of our method for research purposes~\cite{people-snapshot}.


\section{Related Work}
\label{sec:related-work}
Shape reconstruction of humans in clothing can be classified according to two criteria: (1) the type of sensor used and (2) the kind of template prior used for reconstruction.  \emph{Free-form} methods typically use multi-view cameras, depth cameras or fusion of sensors and  reconstruct surface geometry quite accurately without using a strong prior on the shape. In more unconstrained and ambiguous settings, such as in the monocular case, a parametric body model helps to constrain the problem significantly. 
Here we review free-form and model-based methods and focus on methods for monocular images. 
\vspace{-3mm}
\paragraph{Free-form} methods reconstruct the moving geometry by deforming a mesh~\cite{carranza2003free,deAguiar2008performance,cagniart_meshdeform} or using a volumetric representation of shape~\cite{huang2016volumetric,InriaVolumetric_2015}. The advantage of these methods is that they allow reconstruction of general dynamic shapes provided that a template surface is available initially. While flexible, such approaches require high-quality \emph{multi-view} input data which makes them impractical for many applications. Only one approach showed reconstruction of human pose and deforming cloth geometry from monocular video using a pre-captured shape template~\cite{xu2017monoperfcap}. 
Using a \emph{depth camera}, systems like KinectFusion~\cite{izadi2011kinectfusion,newcombe2011kinectfusion} allow reconstruction of 3D rigid scenes and also appearance models~\cite{zhou2014color} by incrementally fusing geometry in a canonical frame. A number of methods adapt KinectFusion for human body scanning~\cite{shapiro2014rapid,3Dportraits,zeng2013templateless,cui2012kinectavatar}. The problem is that these methods require separate shots at different time instances. The person thus needs to stand still while the camera is turned around, or subtle pose changes need to be explicitly compensated. The approach in~\cite{newcombe2015dynamicfusion} generalized KinectFusion to non-rigid objects. The approach performs non-rigid registration between the incoming depth frames and a concurrently updated, initially incomplete, template. 
While general, such template-free approaches~\cite{newcombe2011kinectfusion,innmann2016volume,slavcheva2017killingfusion} are limited to slow and careful motions. One way to make fusion and tracking more robust is by using multiple kinects~\cite{dou2016fusion4d,orts2016holoportation} or multi-view~\cite{starck2007surface,inria_2017,collet2015high}; such methods achieve impressive reconstructions but do not register all frames to the same template and focus on different applications such as streaming or remote rendering for telepresence, e.g., in the holoportation project~\cite{orts2016holoportation}. Pre-scanning the object or person to be tracked~\cite{zollhofer2014real,deAguiar2008performance} reduces the problem to tracking the non-rigid deformations. Some works are in-between free-form and model-based methods. In~\cite{gall2009motion,vlasic2008articulated} they pre-scan a template and insert a skeleton and in~\cite{yu2017bodyfusion} they use a skeleton to regularize dynamic fusion. Our work is also related to the seminal work of~\cite{cheung2003shape,cheung2003visual} where they align visual hulls over time to improve shape estimation. In the articulated case, they need to segment and track every body part separately and then merge the information together in a coarse voxel model; more importantly, they need multi-view input. In~\cite{khan2008reconstructing} they compensate for small motions of captured objects by de-blurring occupancy images but no results are shown for moving humans. In~\cite{zhu2017video} they reconstruct the shape of clothed humans in outdoor environments from RGB video, requiring the subject to stand still. All these works use either multi-view systems, depth cameras or do not handle moving humans. In contrast, we use a single RGB video of a moving person, which makes the problem significantly harder as geometry can not be directly unwarped as it is done in depth fusion papers.
\vspace{-3mm}
\paragraph{Model-based.}
Several works leverage a parametric body model for human pose and shape estimation from images~\cite{PonsModelBased}. Early models in computer vision were based on simple primitives~\cite{metaxas1993shape,gavrila1996,plankers2001articulated,sigal2004tracking}. Recent ones are learned from thousands of scans of real people and encode pose, and shape deformations~\cite{anguelov2005scape,hasler2009statistical,smpl2015loper,zuffi2015stitched,pons2015dyna}. Some works reconstruct the body shape from~\emph{depth data} sequences~\cite{weiss2011home,Helten:2013,ye2014real,zhang2014quality,Bogo:ICCV:2015} exploiting the temporal information. Typically, a single shape and multiple poses are optimized to exploit the temporal information. Using~\emph{multi-view} some works have shown performance capture outdoors~\cite{rhodin2016general,Robertini:2016} by leveraging a sum of Gaussians body model~\cite{stoll2011fast} or using a pre-computed template~\cite{Yu_2015_ICCV}.
A number of works are restricted to estimating the shape parameters of a body model~\cite{bualan2008naked,guan2009estimating} from multiple views or single images with manually clicked points; silhouettes shading cues and color have been used for inference. Some works fit a body model to images using manual intervention~\cite{zhou2010parametric,jain2010moviereshape,rogge2014garment} with the goal of image manipulation. Shape and clothing from a single image is recovered in~\cite{guo2012clothed,chen2013deformable} but the user needs to click points in the image and select the clothing types from a database. In~\cite{kraevoy2009modeling} they obtain shape from contour drawings. The advance in 2D pose detection~\cite{wei2016cpm,cao2017realtime,insafutdinov17cvpr} has made 3D pose and shape estimation possible in challenging scenarios. In~\cite{bogo2016smplify,Lassner} they fit a 3D body model~\cite{smpl2015loper} to 2D detections; since only model parameters are optimized and these methods heavily rely on 2D detections, results tend to be close to the shape space mean. In~\cite{alldieck2017optical} they add a silhouette term to reduce this effect.
\vspace{-3mm}
\paragraph{Shape Under Clothing.} The aforementioned methods ignore clothing or treat it as noise, but a number of works explicitly reason about clothing. Typically, these methods incorporate constraints such as the body should lie inside the clothing silhouette. 
In~\cite{bualan2008naked} they estimate body shape under clothing by optimizing model parameters for a set of images of the same person in different clothing. In~\cite{wuhrer2014estimation,yang2016estimation} they exploit temporal sequences of scans to estimate shape under clothing. Results are usually restricted to the (naked) model space. In~\cite{shape_under_cloth:CVPR17} they estimate detailed shape under clothing from scan sequences by optimizing a free-form surface constrained by a body model. The approach in~\cite{Pons-Moll:Siggraph2017} jointly captures clothing geometry and body shape using separate meshes but requires 3D scan sequences as input. DoubleFusion~\cite{DoubleFusion2018} reconstructs clothing geometry and inner body shape from a single depth camera in real time.
\vspace{-3mm}
\paragraph{Learning based.} Only very few works predict human shape from images using learning methods since images annotated with ground truth shape, pose and clothing geometry are hardly available. A few exceptions are the approach of~\cite{dibra2017human} that predicts shape from silhouettes using a neural network and~\cite{danvevrek2017deepgarment} that predicts garment geometry from a single image. Predictions in~\cite{dibra2017human} are restricted to model shape space and tend to look over-smooth; only garments seen in the dataset can be recovered in~\cite{danvevrek2017deepgarment}. Recent works leverage 2D annotations to train networks for the task of 3D pose estimation~\cite{mehta2017vnect,popa2017deep, zhou2017towards,sun2017compositional,tome2017lifting,rogez_lcr_cvpr17}. Such works typically predict a stick figure or bone skeleton only, and can not estimate body shape or clothing.


\section{Method}
\label{sec:method}
\vspace{-1mm}
Given a single monocular RGB video depicting a moving person, our goal is to generate a personalized 3D model of the subject, which consists of the shape of body, hair and clothing, a personalized texture map, and an underlying skeleton rigged to the surface. Non-rigid surface deformations in new poses are thus entirely skeleton-driven.
Our method consists of 3 steps: 1) \emph{pose reconstruction} (Sec.~\ref{subsec:pose_reconstruction}) 2) \emph{consensus shape estimation} (Sec.~\ref{subsec:consensus_shape}) and 3) \emph{frame refinement and texture map generation} (Sec.~\ref{subsec:texture}). Our main contribution is step 2), the consensus shape estimation; step 1) builds on previous work and step 3) to obtain texture and time-varying details is optional.

In order to estimate the consensus shape of the subject, we first calculate the 3D pose in each frame (Sec.~\ref{subsec:pose_reconstruction}). We extend the method of~ \cite{bogo2016smplify} to make it more robust and enforce better temporal coherence and silhouette overlap. In the second step, the \emph{consensus shape} is calculated as detailed in Sec.~\ref{subsec:consensus_shape}. The consensus shape is efficiently optimized to maximally explain the silhouettes at each frame instance. Due to time-varying cloth deformations the posed consensus shape might be slightly misaligned with the frame silhouettes. Hence, in order to compute texture and capture time-varying details, in step 3) deviations from the consensus shape are optimized per frame in a sliding window approach (Sec.~\ref{subsec:texture}). Given the refined frame-wise shapes we can compute the texture map.
Our method relies on a foreground segmentation of the images. Therefore, we adopt the CNN based video segmentation method of~\cite{Cae+17} and train it with 3-4 manual segmentations per sequence.
In order to counter ambiguities in monocular 3D human shape reconstruction, we use the SMPL body model~\cite{smpl2015loper} as starting point. In the following, we briefly explain how we adapt original SMPL body model for our problem formulation.
\subsection{SMPL Body Model with Offsets}
\label{subsec:body_model}
SMPL is a parameterized model of naked humans that takes $72$ pose and $10$ shape parameters and returns a triangulated mesh with $N=6890$ vertices. The shape $\shape$ and pose $\pose$ deformations are applied to a base template $\template$, which in the original SMPL model corresponds to the statistical mean shape in the training scans $\templatetpose$:
\begin{equation}
\smpl(\shape,\pose) = \blendfun(\posefun(\shape,\pose), \jointfun(\shape), \pose, \blendweights)
\end{equation}
\begin{equation}
\posefun(\shape,\pose) = \templatetpose + \offsetfun_s(\shape) + \offsetfun_p(\pose)
\end{equation}
where $\blendfun$ is a linear blend-skinning function 
applied to a rest pose $\posefun(\shape,\pose)$ based on the skeleton joints $\jointfun(\shape)$  and after pose-dependent deformations $\offsetfun_p(\pose)$ and shape dependent deformations $\offsetfun_p(\pose)$ are applied. Shape-dependent deformations $\offsetfun_s(\shape)$ model subject identity. However the 
Principal Component shape space of SMPL was learned from scans of naked humans,
so clothing and other personal surface detail cannot be modeled. 
In order to personalize the SMPL model, we simply add a set of auxiliary variables or offsets $\offsets \in \mathbb{R}^{3N}$ from the template:
\begin{equation}
\posefun(\pose,\shape,\offsets) = \templatetpose + \offsetfun_s(\shape) + \offsetfun_p(\pose)+\offsets
\label{eq:offset_SMPL}
\end{equation}
Such offsets $\offsets$ allow us to deform the model to better explain details and clothing. Offsets are optimized in step 2. 
\subsection{Pose Reconstruction} 
\label{subsec:pose_reconstruction}
The approach in~\cite{bogo2016smplify} optimizes SMPL model parameters to fit a set of 2D joint detections in the image. As with any monocular method, scale is an inherent ambiguity. To mitigate this effect, we take inspiration from~\cite{rhodin2016general}
and extend~\cite{bogo2016smplify} such that it jointly considers $P=5$ frames and optimizes a single shape and $P=5$ poses. Note that optimizing many more frames would become computationally very expensive and many models would have to be simultaneously stored in memory.
Our experiments reveal that even when optimizing over $P=5$ poses the scale ambiguity prevails. The reason is that pose differences induce 
additional 3D ambiguities which cannot be uniquely decoupled from global size, even on multiple frames
~\cite{taylor2000reconstruction,sminchisescu2003kinematic,PonsMoll_CVPR2014}. Hence, if the height of the person is known, we incorporate it as constraint during optimization. If height is not known the shape reconstructions of our method are still accurate up to a scale factor (height estimation is roughly off by 2-5 cm). The output of initialization are SMPL model shape parameters $\shape_0$ that we keep fixed during subsequent frame-wise pose estimation. In order to estimate 3D pose more reliably, we extend~\cite{bogo2016smplify} by incorporating a silhouette term:
\begin{equation}
E_{\text{silh}}(\pose) = G(\mathbf{w}_{\text{o}}\mathbf{I}_{rn}(\pose)\mathbf{C} + \mathbf{w}_{\text{i}}(\mathbf{1} - \mathbf{I}_{rn}(\pose))\bar{\mathbf{C}} )
\end{equation}
with the silhouette image of the rendered model $\mathbf{I}_{rn}(\pose)$, distance transform of observed image mask $\mathbf{C}$ and its inverse $\bar{\mat{C}}$, weights $\mathbf{w}$. To be robust to local minima we optimize at 4 different levels of a Gaussian pyramid $G$. We further update the method to use state of the art 2D joint detections \cite{cao2017realtime, wei2016cpm} and a single-modal A-pose prior. We train the prior from SMPL poses fitted against body scans of people in A-pose. Further, we enforce a temporal smoothness and initialize the pose in a new frame with the estimated pose $\pose$ in the previous frame.  If the objective error gets too large, we re-initialize the tracker by setting the pose to zero. While optimization in batches of frames would be beneficial it slows down computation and we have not found significant differences in pose accuracy. The output of this step is a set of poses $\{\pose_p\}_{p=1}^F$ for the $F$ frames in the sequence. 
\begin{figure}
	\includegraphics[width=\columnwidth]{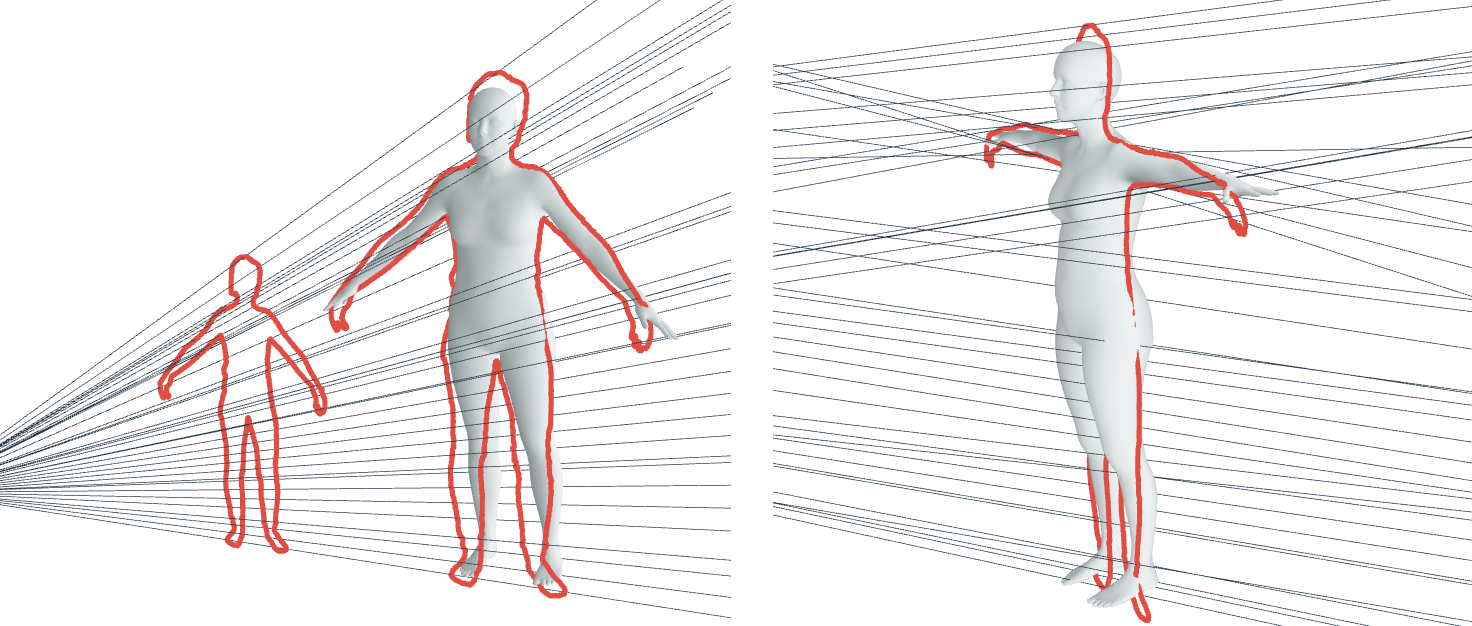}
	\caption{The camera rays that form the image silhouette (left) are getting unposed into the canonical T-pose (right). This allows efficient shape optimization on a single model for multiple frames.}
	\label{fig:unpose_rays}
	\vspace{-.4cm}
\end{figure}
\subsection{Consensus Shape}
\label{subsec:consensus_shape}
Given the set of estimated poses we could jointly optimize a single refined shape matching all original $F$ poses, which would yield a complex, non-convex optimization problem. Instead, we merge all the information into an unposed canonical frame, where refinement is computationally easier. At every frame a silhouette places a new constraint on the body shape; specifically, the set of rays going from the camera to the silhouette points define a constraint cone, see Fig.~\ref{fig:unpose_rays}. Since the person is moving, the pose is changing. Our key idea is to \emph{unpose} the cone defined by the projection rays using the estimated poses. Effectively, we invert the SMPL function for every ray. In SMPL, every vertex $\vertex$ deforms according to the following equation:
\begin{equation}
\small
\vertex_i^{\prime} = \sum_{k=1}^{K}\blendweight_{k,i}\jointtransfun_k(\pose,\jointfun(\shape))(\vertex_i + \offsetfunvert_{s,i}(\shape) + \offsetfunvert_{P,i}(\pose))
\label{eq:SMPL_vert}
\end{equation}
where $\jointtransfun_k$ is the global transformation of joint $k$ and $b_{s,i}(\shape)\in\mathbb{R}$ and $\offsetfunvert_{P,i}(\pose)$ are elements of $\offsetfun_s(\shape)$ and $\offsetfun_p(\pose)$ corresponding to $i-th$ vertex. For every ray $\ray$ we find its closest 3D model point. From Eq.~\eqref{eq:SMPL_vert} it follows that the inverse transformation applied to a ray $\ray$ corresponding to model point $\vertex_i^\prime$ is
\begin{equation}
\small
\ray = \left(\sum_{k=1}^{K}\blendweight_{k,i}\jointtransfun_k(\pose,\jointfun(\shape))\right)^{-1}\ray^{\prime}- \offsetfunvert_{P,i}(\pose).
\end{equation}
Doing this for every ray effectively unposes the silhouette cone and places constraints on a canonical T-pose, see Fig.~\ref{fig:unpose_rays}. Unposing removes blend-shape calculations from the optimization problem and significantly reduces the memory foot-print of the method. Without unposing the vertex operations and the respective Jacobians would have to be computed for every frame at \emph{every update} of the shape.
Given the set of unposed rays for $F$ silhouettes (we use $F=120$ in all experiments), we formulate an optimization in the canonical frame
\begin{equation}
E_{\text{cons}} = E_{\text{data}} + w_{\text{lp}}E_{\text{lp}} + w_{\text{var}}E_{\text{var}} + w_{\text{sym}}E_{\text{sym}}
\label{eq:energy_consensus}
\end{equation}
and minimize it with respect to shape parameters $\shape$ of a template model and the vertex offsets $\offsets$ defined in Eq.~\ref{eq:offset_SMPL}. The objective $E_{\text{cons}}$ consists of a data term $E_{\text{data}}$ and three regularization terms $E_{\text{lp}},E_{\text{var}},E_{\text{sym}}$ with weights $w_*$ that balance its influence.
\paragraph{Data Term} measures the distance between vertices and rays. Point to line distances can be efficiently computed expressing rays using Plucker coordinates ($\ray=\ray_m,\ray_n)$. Given a set of correspondences $(\vertex_i,\ray)\in\mathcal{M}$ the data term equals
\begin{equation}
E_{\text{data}} = \sum_{(\vertex,\ray) \in \mathcal{M}} \rho(\vertex \times \ray_n - \ray_m)
\end{equation}
where $\rho$ is the Geman-McClure robust cost function, here applied to the point to line distance. Since the canonical pose parameters are all zero ($\pose = \mathbf{0}$) it follows from Eq.~\ref{eq:offset_SMPL} that vertex positions are a function of shape parameters and offsets $\vertex(\shape_0,\mathbf{D}) = T_i(\shape_0,\mathbf{D}) = (\vertex_{\mu,i} + \offsetfunvert_{s,i}(\shape_0) +\mathbf{d}_i)$, where $\mathbf{d}_i \in \mathbb{R}^3$ is the offset in $\offsets$ corresponding to vertex $\vertex_i$. In our notation, we remove the dependency on parameters for clarity.  
The remaining terms regularize the optimization.
\paragraph{Laplacian Term.} We enforce smooth deformation by adding the Laplacian mesh regularizer~\cite{sorkine2004laplacian}:
\begin{equation}
\small
E_{\text{lp}} = \sum_{i=1}^{N} \vertexweight_{l,i} || L(\vertex_i) - \delta_i ||^2
\end{equation}
where $\delta = L(\vertex(\shape_0,\mathbf{0}))$ and $L$ is the Laplace operator. The term forces the Laplacian of the optimized mesh to be similar to the Laplacian of the mesh at initialization (where offsets $\mathbf{D} = \mathbf{0}$).
\paragraph{Body Model Term.} 
We penalize deviations of the reconstructed free-form vertices $\vertex(\shape_0,\mathbf{D})$ from vertices explained by the SMPL model $\vertex(\shape, \mathbf{0})$: 
\begin{equation}
\small
E_{\text{var}} = \sum_{i=1}^{N} \vertexweight_{v,i} ||  \vertex_i(\shape_0,\mathbf{D}) - \vertex_i(\shape,\mathbf{0})||^2
\end{equation}
\paragraph{Symmetry Term.}
Humans are usually axially symmetrical with respect to the Y-axis. Since the body model is nearly symmetric, we add a constraint on the offsets alone that enforces a symmetrical shape:
\begin{equation}
\small
E_{\text{sym}} = \sum_{(i,j)\in \mathcal{S}} \vertexweight_{s,i,j} \left|\left| [-1, 1, 1]^T \cdot \mathbf{d}_i - \mathbf{d}_j \right|\right|^2
\end{equation}
where $\mathcal{S}$ contains all pairs of Y-symmetric vertices. We phrase this as a soft-constraint to allow potential asymmetries in clothing wrinkles and body shapes. 
Since the refined consensus shape still has the mesh topology of SMPL, we can apply the pose-based deformation space of SMPL to simulate surface deformation in new skeleton poses. 
\paragraph{Implementation Details.}
Body regions that are typically unclothed or where silhouettes are noisy (face, ears, hands, and feet) are more regularized towards the body model using per-vertex weights $\vec{\vertexweight}$. We optimize $E_{\text{cons}}$ using a ``dog-leg'' trust region method using the chumpy auto-differentiation framework. We alternate minimizing $E_{\text{cons}}$ with respect to model parameters and offsets and finding point to line correspondences. We also re-initialize $E_{\text{lp}}$, $E_{\text{var}}$, $E_{\text{sym}}$.
More implementation details and runtime metrics are given in the supplementary material.
\vspace{1mm}
\subsection{Frame Refinement and Texture Generation}
\label{subsec:texture}
\begin{figure}
	\centering
	\begin{minipage}{0.39\columnwidth}\centering\offinterlineskip
		\includegraphics[width=0.33\columnwidth]{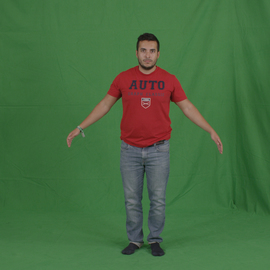}%
		\includegraphics[width=0.33\columnwidth]{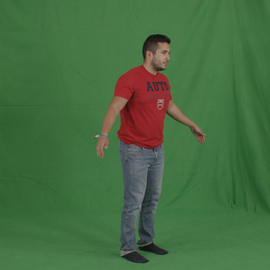}%
		\includegraphics[width=0.33\columnwidth]{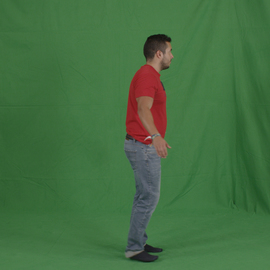}\\
		\includegraphics[width=0.33\columnwidth]{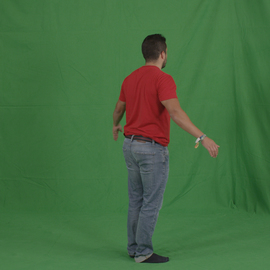}%
		\includegraphics[width=0.33\columnwidth]{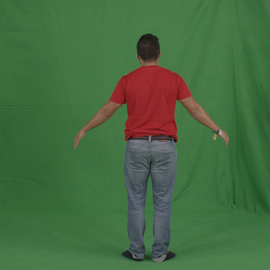}%
		\includegraphics[width=0.33\columnwidth]{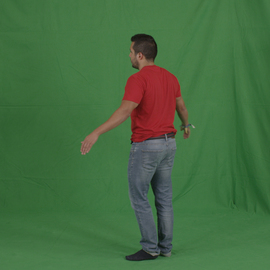}\\
		\includegraphics[width=0.33\columnwidth]{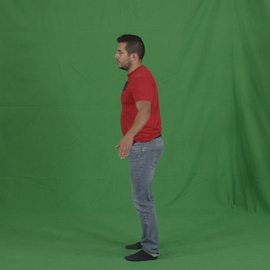}%
		\includegraphics[width=0.33\columnwidth]{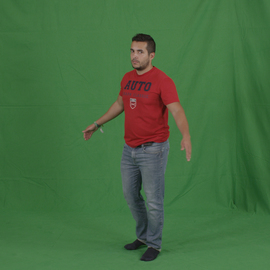}%
		\includegraphics[width=0.33\columnwidth]{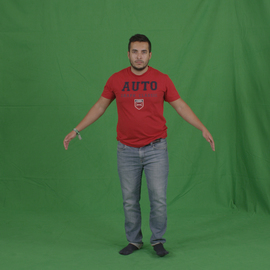}%
	\end{minipage}%
	\begin{minipage}{0.39\columnwidth}\centering
		\includegraphics[width=\columnwidth]{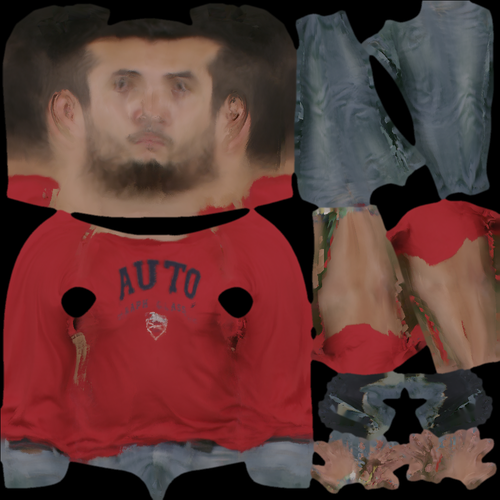}
	\end{minipage}
	\vspace{1mm}
	\caption{We back-project the image color from several frames to all visible vertices to generate a full texture map.}
	\label{fig:texture}
	\vspace{-4mm}
\end{figure}

After calculating a \emph{global} shape for the given sequence, we aim to capture the temporal variations. 
We adapt the energy in Eq.~\ref{eq:energy_consensus} to process frames sequentially. The optimization is initialized with the preceding frame and regularized with neighboring frames:
\begin{align}
\small
E_{\text{ref},j} = \sum_{j=f-m}^{f+m} \psi_j E_{\text{data},j} + w_{\text{var}}E_{\text{var},j} \nonumber \\ + w_{\text{lp}}E_{\text{lp},j} + w_{\text{last}}E_{\text{last},j}
\end{align}
where $\psi_j =1$ for $j=k$ and $\psi_j= w_{\text{neigh}}<1$ for neighboring frames. Hence, $w_{\text{neigh}}$ defines the influence of neighboring frames and $E_{\text{last}}$ regularizes the reconstruction to the result of the preceding frame. To create the texture, we warp our estimated canonical model back to each frame, back-project the image color to all visible vertices, and finally generate a texture image by calculating the median of the most orthogonal texels from all views. An example of keyframes we use for texture mapping and the resulting texture image is shown in Fig.~\ref{fig:texture}.

\section{Experiments}
\vspace{-1mm}
We study the effectiveness of our method, qualitatively and quantitatively, in different scenarios.
For quantitative evaluation, we used two publicly available datasets consisting of 3D scan sequences of humans in motion: with minimal clothing (MC) (DynamicFAUST~\cite{dfaust:CVPR:2017}) and with clothing (BUFF ~\cite{shape_under_cloth:CVPR17}). Since these datasets were recorded without RGB sensors we simply render images of the scans using a virtual camera and use them as input.   
In order to evaluate our method on more varied clothing and backgrounds, we captured a new test dataset (People-Snapshot dataset), and present qualitative results.
To the best of our knowledge, our method is the first approach that enables detailed human body model reconstruction in clothing from a single monocular RGB video without requiring a pre-scanned template or manually clicked points.
Thus, there exist no methods with the same setting as ours. Hence, we provide a quantitative comparison to the state-of-the-art RGB-D based approach KinectCap~\cite{Bogo:ICCV:2015} on their dataset. 
The image sequences and ground truth scans were provided by the authors of~\cite{Bogo:ICCV:2015}. 
While reconstruction from monocular videos is much harder than from depth videos, a comparison is still informative. 
In all experiments, the method's parameters are set to two constant values, one set for clothed and one set for people in MC, which are empirically determined.

\vspace{-0.5mm}
\subsection{Results on Rendered Images}
\vspace{-0.5mm}
We take all 9 sequences of 5 different subjects in the BUFF dataset and all 9 sequences of 9 subjects from the DynamicFaust dataset performing ``Hip'' movements, featuring strong fabric movement or soft tissue dynamics respectively.
Each dynamic sequence consists of 300-800 frames. To simulate the subject rotating in front of a camera, we create a virtual camera at 2.5 meters away from the 3D scans of the subject. We rotate the camera in a circle around the person moving one time per sequence.
The foreground masks are easily obtained from the alpha channel of the rendered images.
For BUFF we render images with real dynamic textures; for DynamicFAUST since textures are not available we rendered shaded models.
\begin{figure}[t]
	\begin{center}\offinterlineskip
		\includegraphics[width=0.33\columnwidth]{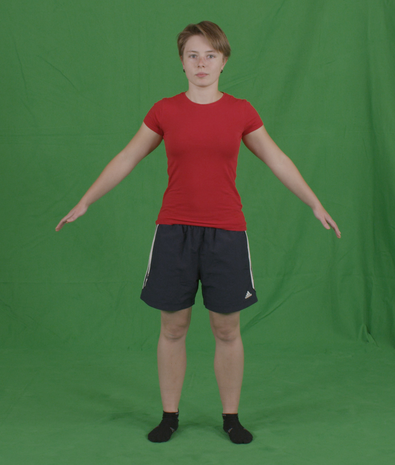}%
		\includegraphics[width=0.33\columnwidth]{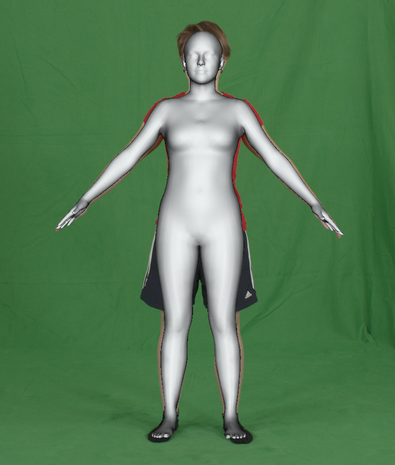}%
		\includegraphics[width=0.33\columnwidth]{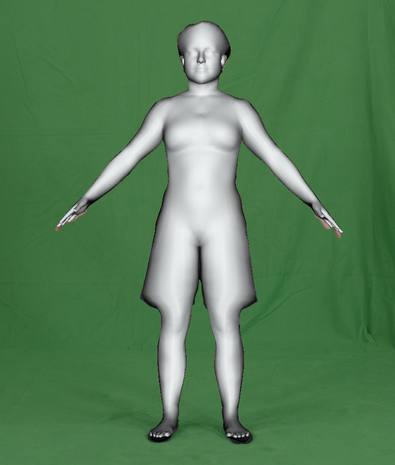}\\
		\vspace{-3mm}
	\end{center}
	\caption{Comparison to the monocular model-based method~\cite{bogo2016smplify} (left to right) input frame, SMPLify, consensus shape. To make a fair comparison we extended~\cite{bogo2016smplify} to multiple views as well. Compared to pure model-based methods, our approach captures also medium level geometry details from a single RGB camera.}
	\label{fig:SMPLify}
	\vspace{-5mm}
\end{figure}
\begin{figure*}[t]
	\hspace{0.02\textwidth}%
	\begin{minipage}[t]{0.52\textwidth}\centering\offinterlineskip
		\includegraphics[height=0.32\columnwidth]{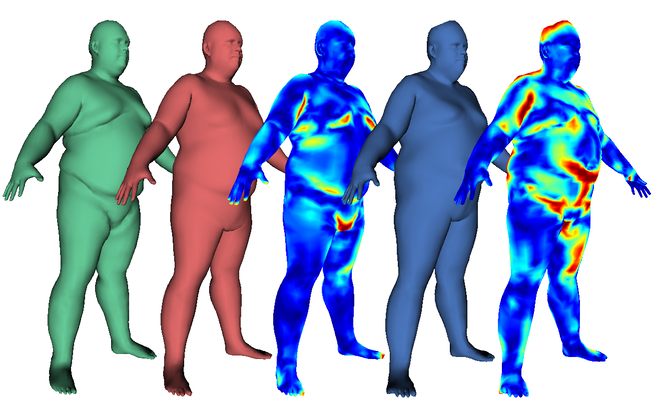}~
		\includegraphics[height=0.32\columnwidth]{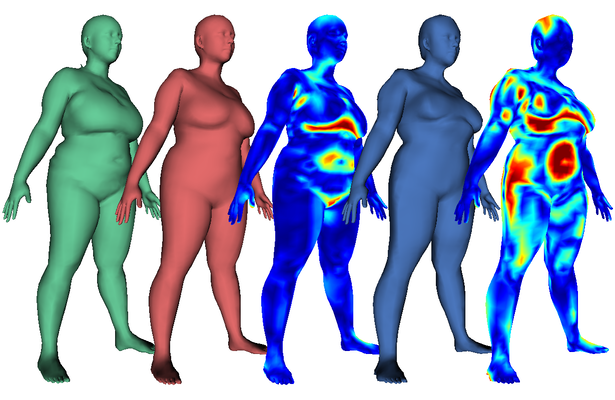}\\
		\includegraphics[height=0.32\columnwidth]{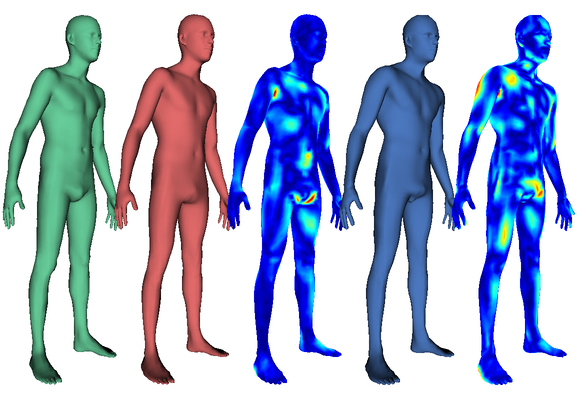}~
		\includegraphics[height=0.32\columnwidth]{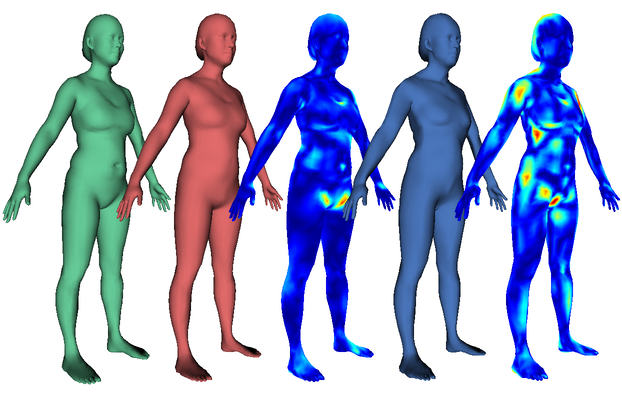}\\
		{\footnotesize
			\begin{minipage}[t]{\textwidth}
				\vspace{-2mm}
				\hspace{7mm}a)
				\hspace{5.5mm}b)
				\hspace{5.5mm}c)
				\hspace{5.5mm}d)
				\hspace{5.5mm}e)
			\end{minipage}%
		}
	\end{minipage}
	\hspace{0.04\textwidth}%
	\begin{minipage}[t]{0.37\textwidth}\centering\offinterlineskip
		\includegraphics[height=0.44\columnwidth]{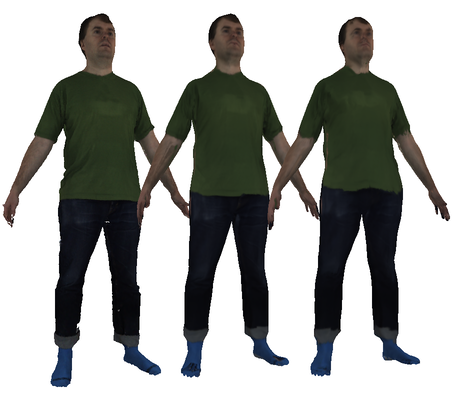}~
		\includegraphics[height=0.44\columnwidth]{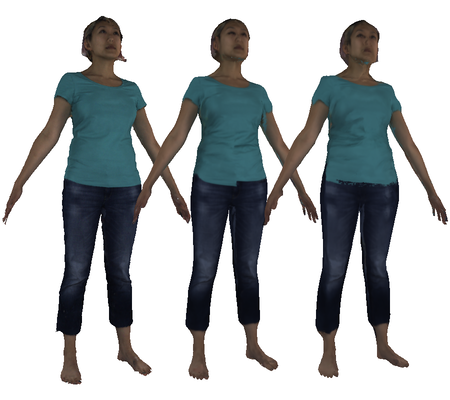}\\
		\includegraphics[height=0.44\columnwidth]{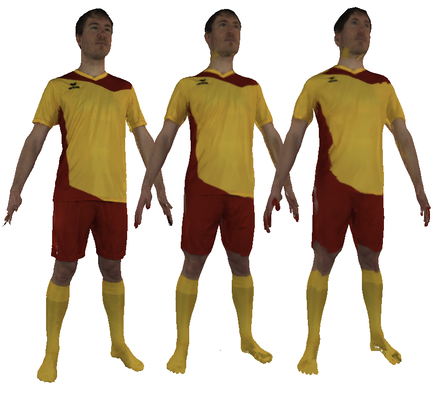}~
		\includegraphics[height=0.44\columnwidth]{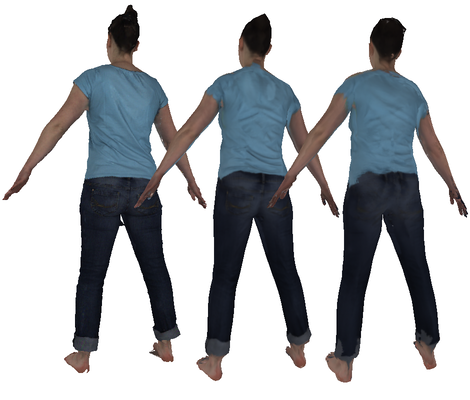}\\
		{\footnotesize
			\begin{minipage}[t]{\textwidth}
				\vspace{-2mm}
				\hspace{6mm}a)
				\hspace{6mm}b)
				\hspace{6mm}c)
			\end{minipage}%
		}
	\end{minipage}
	\vspace{1mm}
	\caption{
		Our results on image sequences from BUFF and D-FAUST datasets. Left we show D-FAUST: (a) ground truth 3D scan, (b) consensus shape with ground truth poses (consensus-p), (c) consensus-p heatmap, (d) consensus shape (consensus), (e) consensus heat-map (blue means 0mm, red means $\geq$ 2cm). Right we show textured results on BUFF: (a) ground truth scan, (b) consensus-p (c)  consensus.
	}	
	\label{fig:buff}
\end{figure*}
\begin{table*}
	\vspace{-2.2mm}
	\begin{minipage}[t]{0.3\textwidth}\centering
		{\scriptsize
			
			\begin{tabular}[t]{l|c||c} 
				\hline 
				\emph{Subject ID} & \emph{full method} & \emph{GT poses}\\ 
				\hline 
				50002 & 5.13 $\pm$6.43 & 3.92 $\pm$4.49 \\
				50004 & 4.36 $\pm$4.67 & 2.95 $\pm$3.11 \\
				50009 & 3.72 $\pm$3.76 & 2.56 $\pm$2.50 \\
				50020 & 3.32 $\pm$3.04 & 2.27 $\pm$2.06 \\
				50021 & 4.45 $\pm$4.05 & 3.00 $\pm$2.66 \\
				50022 & 5.71 $\pm$5.78 & 2.96 $\pm$2.97 \\
				50025 & 4.84 $\pm$4.75 & 2.92 $\pm$2.94 \\
				50026 & 4.56 $\pm$4.83 & 2.62 $\pm$2.48 \\
				50027 & 3.89 $\pm$3.57 & 2.55 $\pm$2.33 \\
				\hline\end{tabular}
		} 
	\end{minipage}%
	\begin{minipage}[t]{0.32\textwidth}\centering
		{\scriptsize
			\begin{tabular}[t]{l|cc||c} 
				\hline 
				\multicolumn{2}{l}{\emph{Subject ID}} & \emph{full method} & \emph{GT poses} \\ 
				\hline 
				\parbox[t]{2mm}{\multirow{5}{*}{\rotatebox[origin=c]{90}{{\tiny t-shirt, long pants}}}} & 00005  & 5.07 $\pm$5.74 & 3.80 $\pm$4.13 \\
				&00032 & 4.84 $\pm$5.25 & 3.37 $\pm$3.59 \\
				&00096 & 5.57 $\pm$6.54 & 4.35 $\pm$4.66 \\
				&00114  & 4.22 $\pm$5.12 & 3.14 $\pm$2.99 \\
				&03223 & 4.85 $\pm$4.80 & 2.87 $\pm$2.58 \\
				\hline 
				\parbox[t]{2mm}{\multirow{4}{*}{\rotatebox[origin=c]{90}{{\tiny  soccer outfit}}}} & 00005 & 5.35 $\pm$6.67 & 3.82 $\pm$3.67 \\
				&00032 & 7.95 $\pm$8.62 & 3.04 $\pm$3.39 \\
				&00114 & 4.97 $\pm$5.81 & 3.01 $\pm$2.80 \\
				&03223 & 5.49 $\pm$5.71 & 3.21 $\pm$3.28 \\
				\hline\end{tabular}
		} 
	\end{minipage}%
	\begin{minipage}[t]{0.38\textwidth}\centering
		{\scriptsize
			\vspace{5mm}
			\begin{tabular}[t]{l|c||l|c} 
				\hline 
				\emph{Subject ID} & & \emph{Subject ID}  & \\ 
				\hline 
				00009 & 4.07 $\pm$4.20 & 02909 & 3.94 $\pm$4.80 \\
				00043 & 4.30 $\pm$4.39 & 03122 & 3.21 $\pm$2.85 \\
				00059 & 3.87 $\pm$3.96 & 03123 & 3.68 $\pm$3.22 \\
				00114 & 4.85 $\pm$4.93 & 03124 & 3.67 $\pm$3.31 \\
				00118 & 3.79 $\pm$3.80 & 03126 & 4.89 $\pm$6.12 \\
				
				\hline\end{tabular}
		} 
	\end{minipage}
	\vspace{-0.5mm}
	\caption{Numerical evaluation on 3 different datasets with ground truth 3D shapes. On D-FAUST and BUFF we rendered the ground truth scans on a virtual camera (see text), KinectCap already included images. We report for every subject the average surface to surface distance (see text). On BUFF, D-FAUST and KinectCap we achieve mean average errors of 5.37mm,  4.44mm,  3.97mm respectively. As expected best results are obtained using ground truth poses. Perhaps surprisingly, the results (3.40 mm for BUFF, 2.86 for D-FAUST) do not differ much from the average errors of the full pipeline. 
		This demonstrates that our approach is robust to inaccuracies in 3D pose estimation. 
	}
	\label{tab:buff_error}
	\vspace{-4mm}
\end{table*}
\begin{figure*}[t]
	\begin{center}
		\includegraphics[width=\linewidth]{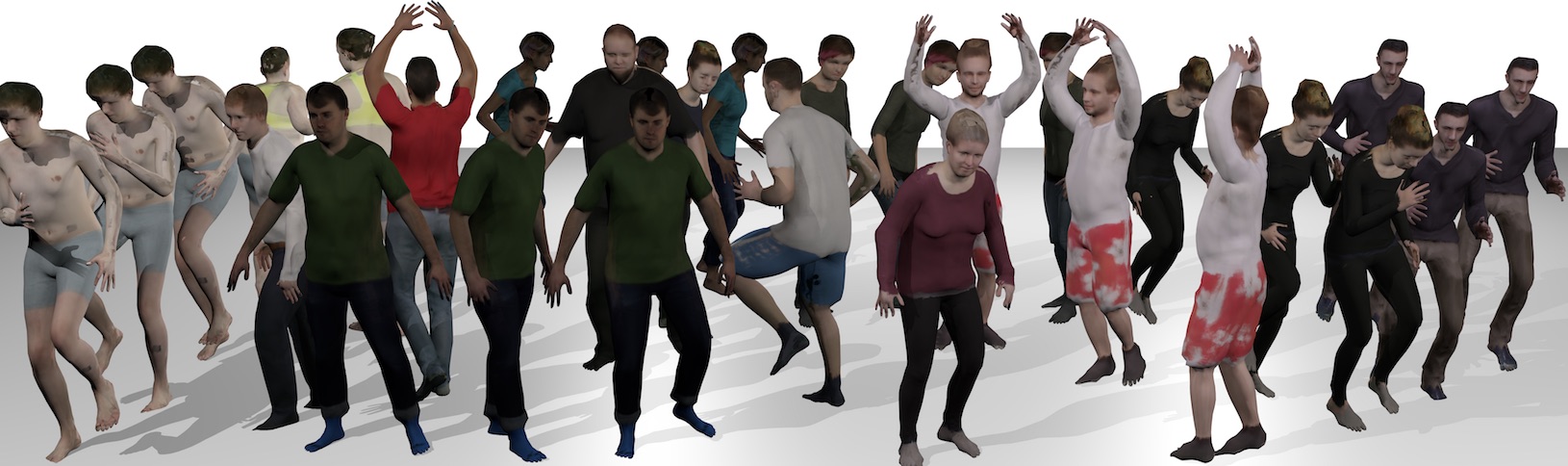}
	\end{center}
	\vspace{-0.3cm}
	\caption{Qualitative results: since the reconstructed templates share the topology with the SMPL body model we can use SMPL to change the \emph{pose and shape} of our reconstructions. While SMPL does not model clothing deformations the deformed templates look plausible and maybe of sufficient quality for several applications.}
	\vspace{-0.3cm}
	\label{fig:shapes_poses}
\end{figure*}
In Fig.~\ref{fig:buff}, we show some examples of our reconstruction results on image sequences rendered from BUFF and DynamicFAUST scans.
The complete results of all 9 sequences are provided in the supplementary material.
To be able to quantitatively evaluate the reconstruction quality, we adjust the pose and scale of our reconstruction to match the ground truth body scans following~\cite{shape_under_cloth:CVPR17, Bogo:ICCV:2015}. Then, we compute a bi-directional vertex to surface distance between our reconstruction and the ground truth geometry.
Per-vertex errors (in millimeters) on all sequences are provided in Tab.~\ref{tab:buff_error}. The heatmaps of per-vertex errors are shown in Fig.~\ref{fig:buff}.
As can be seen, our method yields accurate reconstruction on all sequences including personalized details.
To study the importance of the pose estimation component, we report the accuracy of our method using \emph{ground truth poses} versus using estimated poses \emph{full method}.
Ground truth poses were obtained by registering SMPL to the 3D scans.
The results of the ablation evaluation are also shown in Fig.~\ref{fig:buff} and Tab.~\ref{tab:buff_error}.
We can see that our complete pipeline achieved comparable accuracy with the one using ground truth poses which demonstrates robustness. Results show that there is still room for improvement in 3D pose reconstruction.

\vspace{-1.5mm}
\subsection{Qualitative Results on RGB Images}
\begin{figure}\centering\offinterlineskip
	\includegraphics[height=0.33\columnwidth]{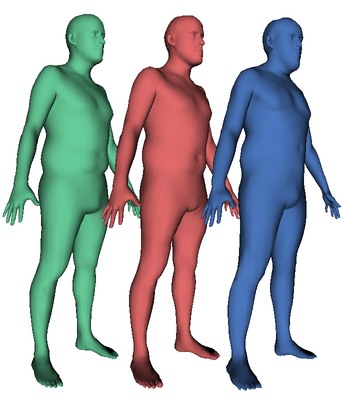}
	\includegraphics[height=0.33\columnwidth]{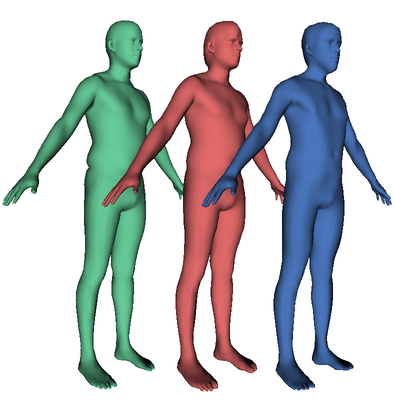}
	\includegraphics[height=0.33\columnwidth]{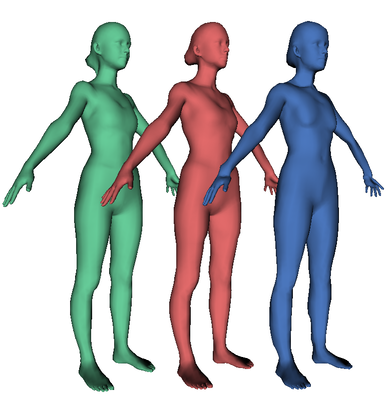}
	\vspace{1mm}
	\caption{Comparison to the RGB-D based method of~\cite{Bogo:ICCV:2015} (red) and ground truth scans (green). Our approach (blue) achieves similar qualitative results despite using a monocular video sequence as opposed to a depth camera. Their approach is more accurate numerically 2.54~mm versus 3.97~mm but our results are comparable despite using a single RGB camera. 
	}
	\vspace{-1.5mm}
	\label{fig:kinect_cap}
\end{figure}
\begin{figure}\centering\offinterlineskip
	\includegraphics[width=0.5\columnwidth]{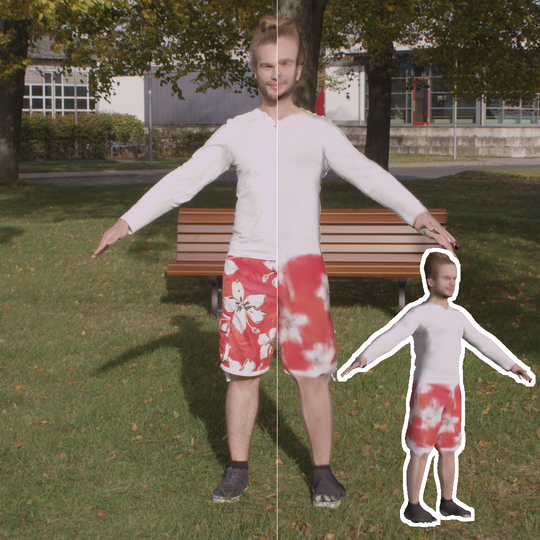}%
	\includegraphics[width=0.5\columnwidth]{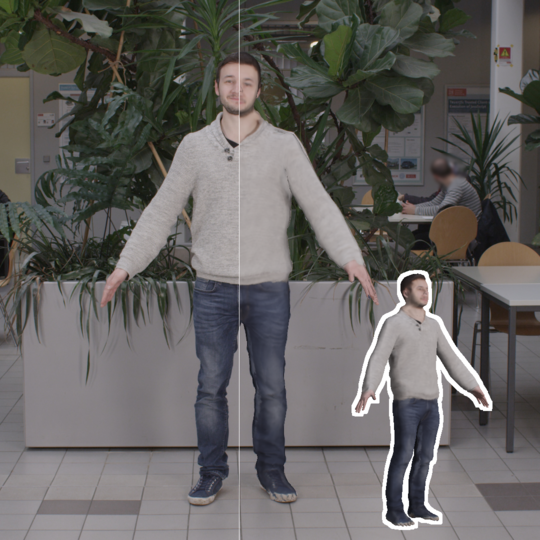}\\
	\includegraphics[width=0.5\columnwidth]{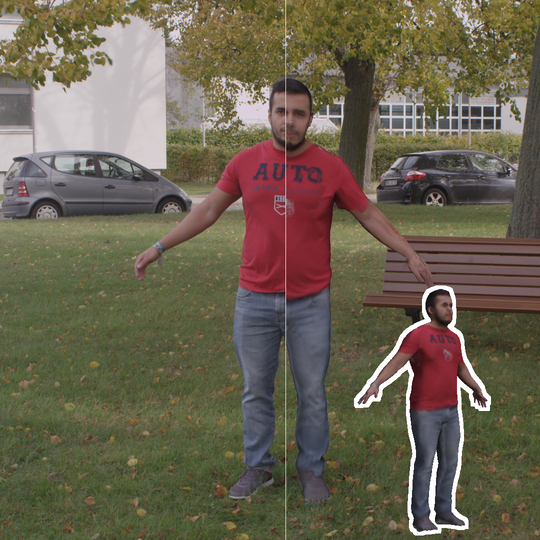}%
	\includegraphics[width=0.5\columnwidth]{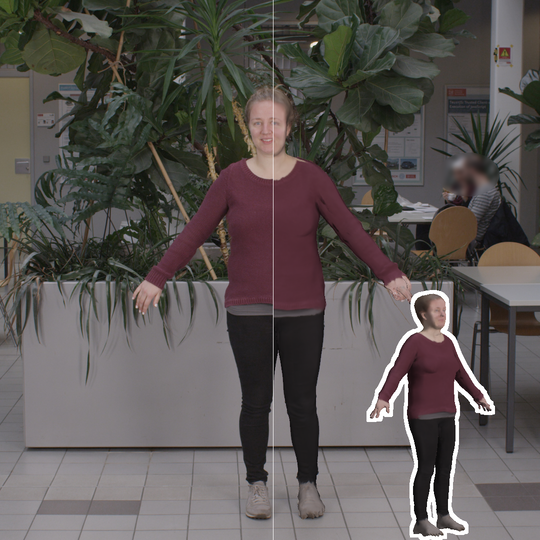}
	\vspace{2mm}
	\caption{Side-by-side comparison of our reconstructions (right) and the input images (left). As can be seen from the right side, our reconstructions precisely overlay on the input images. The reconstructed models rendered in a side view are shown at bottom right.}
	\label{fig:sidebyside}
	\vspace{-3mm}
\end{figure}

We also evaluate our method on real image sequences. The People-Snapshot dataset consists of 24 sequences of $11$ subjects varying a lot in height and weight.
The sequences are captured with a fixed camera, and we ask the subjects to rotate while holding an A-pose.
To cover a variety of clothing, lighting conditions and background, the subjects were captured with varying sets of garments and with three different background scenes: in the studio with green screen, outdoor, and indoor with complex dynamic background.
Some examples of our reconstruction results are shown in Fig.~\ref{fig:shapes_poses} and Fig.~\ref{fig:teaser}. We show more example in the supplementary material and in the video.
We can see that our method yields detailed reconstructions of similar quality as the results on rendered sequences, which demonstrates that our method generalizes well on the real world scenarios. The benefits of our method are further evidenced by overlaying the re-posed final reconstruction on to the input images.
As shown in Fig.~\ref{fig:sidebyside}, our reconstructions precisely overlay the body silhouettes in the input images.
\vspace{-0.5mm}
\subsection{Comparison with KinectCap}
\vspace{-0.5mm}
We compare our method to~\cite{Bogo:ICCV:2015} on their collected dataset.
Subjects were captured in both A-pose and T-poses in this dataset. Since T-poses (zero-pose in SMPL) are rather unnatural, they are not well captured in our general pose-prior. Hence, we adjust our pose prior to contain also T-poses.
Note that their method relies on depth data, while ours only uses the RGB images.
Notably, our method obtains comparable results qualitatively and quantitatively despite solving a much more ill-posed problem.
This is further evidenced by the per-vertex errors in Tab.~\ref{tab:buff_error}.
\vspace{-0.5mm}
\subsection{Surface Refinement Using Shading}
\begin{figure}
	\centering\offinterlineskip
	\includegraphics[width=0.33\columnwidth]{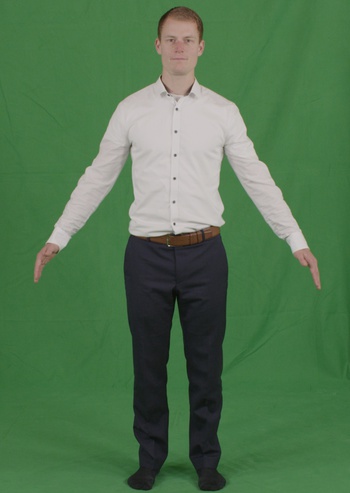}~
	\includegraphics[width=0.33\columnwidth]{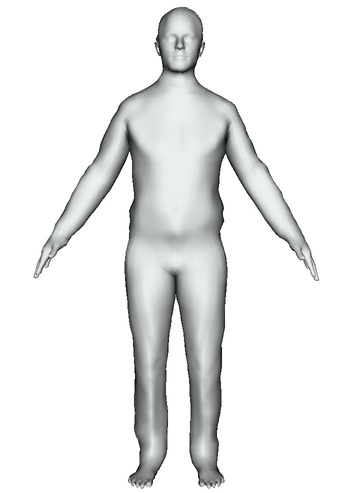}~
	\includegraphics[width=0.33\columnwidth]{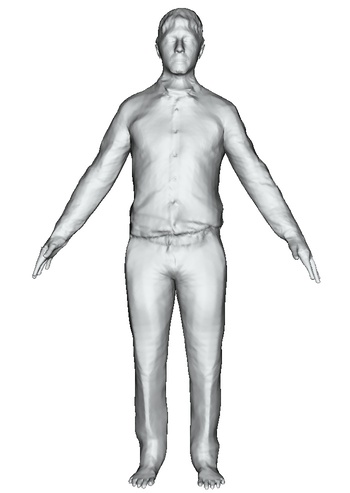}
	\vspace{2mm}
	\caption{Our reconstruction can be further improved by adding the finer level details of the surface using shape from shading.}
	\label{fig:shapefromshading}
\end{figure}

\vspace{-0.5mm}
As mentioned before, our method captures both body shape and medium level surface geometry. In contrast to pure model-based methods, we already add significant details (Fig.~\ref{fig:SMPLify}).
Using existing shape from shading methods the reconstruction can be further improved by adding the finer level details of the surface, e.g. folding and wrinkles.
Fig.~\ref{fig:shapefromshading} shows an example result of applying the shape from shading method of~\cite{sfs:Wu:CVPR2011} to our reconstruction.
This application further demonstrates the accuracy of our reconstruction, since such good result cannot be obtained without an accurate model-to-image alignment.


\section{Discussion and Conclusions}
\label{sec:conclusions}
We have proposed the first approach to reconstruct a personalized 3D human body model from a single video of a moving person. The reconstruction comprises personalized geometry of hair, body, and clothing, surface texture, and an underlying model that allows changes in pose and shape. 
Our approach combines a parametric human body model extended by surface displacements for refinement, and a novel method to morph and fuse the dynamic human silhouette cones in a common frame of reference. The fused cones merge the shape information contained in the video, allowing us to optimize a detailed model shape. 
%
Our algorithm not only captures the geometry and appearance of the surface, but also automatically rigs the body model with a kinematic skeleton enabling approximate pose-dependent surface deformation. Quantitative results demonstrate that our approach can reconstruct human body shape with an accuracy of 4.5mm and an ablation analysis shows robustness to noisy 3D pose estimates. 

The presented method finds its limits in appearances that do not share the same topology as the body: long open hair or skirts can not be modeled as an offset from the body. Furthermore, we can only capture surface details that are seen on the outline of at least one view. This means especially
concave regions like armpits or inner thighs are sometimes not well handled. Strong fabric movement caused by fast skeletal motions will additionally result in decreased level of detail. In future work, we plan to incorporate illumination and material estimation alongside with temporally varying textures in our method to enable realistic rendering and video augmentation.

For the first time, our method can extract realistic avatars including hair and clothing from a moving person in a \emph{monocular RGB video}. Since cameras are ubiquitous and low cost, people will be able to digitize themselves and use the 3D human models for VR applications, entertainment, biometrics or virtual try-on for online shopping. Furthermore, our method precisely aligns models with the images, which opens up many possibilities for image editing.


\vspace{2mm}
\noindent
\textbf{Acknowledgments}\\
The authors gratefully acknowledge funding by the German Science Foundation from project DFG MA2555/12-1. We would like to thank Rudolf Martin and Juan Mateo Castrillon Cuervo for great help in data collection and processing. Another thanks goes to Federica Bogo and Javier Romero for providing their results for comparison.

{\small
\bibliographystyle{ieee}
\bibliography{egbib}
}

\end{document}